\documentclass[conference]{IEEEtran}
\usepackage{graphicx} 
\usepackage{amsmath,amssymb} 
\usepackage{booktabs} 
\usepackage{hyperref}
\hypersetup{colorlinks=true, linkcolor=black, urlcolor=blue, citecolor=black}

\IEEEoverridecommandlockouts
\makeatletter
\def\ps@IEEEtitlepagestyle{%
  \def\@oddfoot{\mycopyrightnotice}%
  \def\@evenfoot{}%
}
\def\mycopyrightnotice{%
  \raisebox{-15pt}[0pt][0pt]{\parbox[b]{\textwidth}{\scriptsize
  \rule{\textwidth}{0.4pt}\\[3pt]
  \copyright~2026 IEEE. Personal use of this material is permitted. Permission from IEEE must be
  obtained for all other uses, in any current or future media, including reprinting/republishing
  this material for advertising or promotional purposes, creating new collective works, for resale
  or redistribution to servers or lists, or reuse of any copyrighted component of this work in
  other works.\\[1pt]
  Accepted version. Version of record: M.~F.~Raza, S.~L.~Yang, and S.~M.~Srinivasan, ``SAGE:
  SLO-Aware Adaptive Retrieval for Production RAG Systems,'' in \emph{2026 12th International
  Conference on Control, Decision and Information Technologies (CoDIT)}, Bari, Italy, 2026,
  pp.~169--175. doi:~10.1109/CoDIT70676.2026.11631166\\[1pt]
  \emph{Author contact: faizanraza766@gmail.com}}}%
  \gdef\mycopyrightnotice{}%
}
\makeatother

\title{SAGE: SLO-Aware Adaptive Retrieval for Production RAG Systems}

\author{%
\begin{tabular*}{\textwidth}{@{\extracolsep{\fill}}ccc}

\parbox[t]{0.31\textwidth}{%
\vspace{0pt}
\centering
\textbf{Muhammad Faizan Raza}\\
\textit{Engineering}\\
\textit{Pennsylvania State University}\\
\textit{Great Valley}\\
Malvern, PA, USA\\
\texttt{mfr5933@psu.edu}\\
{\scriptsize ORCID: 0009-0005-3256-1130}
}
&
\parbox[t]{0.31\textwidth}{%
\vspace{0pt}
\centering
\textbf{Shuo (Luna) Yang}\\
\textit{Business Division}\\
\textit{Pennsylvania State University}\\
\textit{Brandywine}\\
Media, PA, USA\\
\texttt{sfy5287@psu.edu}\\
{\scriptsize ORCID: 0000-0003-2390-243X}
}
&
\parbox[t]{0.31\textwidth}{%
\vspace{0pt}
\centering
\textbf{Satish Mahadevan Srinivasan}\\
\textit{Engineering}\\
\textit{Pennsylvania State University}\\
\textit{Great Valley}\\
Malvern, PA, USA\\
\texttt{sus64@psu.edu}\\
{\scriptsize ORCID: 0000-0003-1377-3726}
}

\end{tabular*}%
}

\begin{document}
\maketitle

\begin{abstract}

Retrieval-Augmented Generation (RAG) systems in production operate under strict service level objectives (SLOs) on tail latency and infrastructure cost. However, standard retrieval pipelines rely on fixed retrieval budgets that ignore query difficulty, over-retrieving for easy queries and under-serving hard ones, forcing operators to trade answer quality against SLO compliance. 

This paper proposes SAGE, a learned \emph{SLO-aware adaptive retrieval policy} that dynamically selects the number of passages \emph{k} per query. SAGE uses lightweight features derived from initial retrieval (e.g., score distributions, rank gaps, lexical signals) and is trained offline via imitation learning from an oracle that approximates optimal latency–quality trade-offs. At inference, it adds no LLM calls and minimal overhead.

On Natural Questions, under a 5\,s P95 latency SLO, SAGE achieves 95\% SLO compliance versus 30\% for the best static baseline ($k{=}20$), reduces P95 latency by 36\% and retrieval cost by 51\% with only 2 percentage points Exact Match (EM) loss. A single policy trained on Natural Questions generalizes across HotpotQA, UnSeenTimeQA, and four LLM families (Llama, Qwen, Mistral, Gemma), consistently yielding +45–52 point SLO improvements without quality degradation.
\end{abstract}


\section{Introduction}
Large language models (LLMs) increasingly rely on \emph{retrieval-augmented generation} (RAG) to reduce hallucinations and incorporate up-to-date knowledge~\cite{lewis2020rag,guu2020realm,karpukhin2020dpr,gao2024rag_survey}. In production, these systems operate under strict  \emph{service level objectives} (SLOs) that constrain not only average latency but also tail latency (e.g., P95 or P99) and infrastructure cost. Meeting such SLOs is critical in user-facing applications such as search and customer support, where slow or inconsistent responses directly impact engagement and operational expenditure~\cite{dean2013tail}.

Most deployed RAG pipelines, however, still rely on a globally tuned, \emph{fixed} retrieval budget $k$ per query. Operators select $k$ empirically (e.g., $k{=}10$ or $20$) to balance answer quality against system load, then apply this choice uniformly across all traffic. This design is fundamentally misaligned with real-world query distributions: easy factoid questions can often be answered from a handful of passages, while hard multi-hop or temporal queries genuinely benefit from deeper retrieval. A single global $k$ inevitably over-spends on easy queries and under-serves hard ones. When $k$ is set high enough to protect answer quality, the resulting retrieval and re-ranking workload drives up tail latency and cost, leading to widespread SLO violations.

Recent work on adaptive and active RAG explores conditioning retrieval on model uncertainty or utility signals, deciding \emph{whether} to retrieve, \emph{when} to stop iterating, or \emph{which} retrieval strategy to invoke~\cite{jiang2023flare,asai2024selfrag,adaptivek2025,stoprag2025}. While these approaches demonstrate that adaptive retrieval can improve factuality and efficiency, they are not directly optimized for production SLOs: many require additional LLM calls, involve complex multi-step protocols, or optimize surrogate objectives that only indirectly reflect latency percentiles and cost. There remains a gap between adaptive RAG algorithms and the needs of operators who must satisfy contractual SLOs on fixed hardware budgets.

This paper addresses that gap by treating retrieval as a \emph{per-query resource allocation decision} under explicit latency and cost constraints. We introduce SAGE, an \emph{SLO-aware adaptive retrieval} policy that predicts, for each query, an appropriate retrieval budget $k$ before generation. SAGE operates entirely on lightweight features already available in standard RAG stacks, BM25 and dense retriever scores, rank statistics, and simple lexical signals, to estimate query difficulty and the marginal value of additional passages. The policy is trained offline using labels derived from budget sweeps under a target P95 SLO.

On Natural Questions, under a 5\,s P95 SLO, SAGE improves SLO compliance from 30\% (best static $k{=}20$) to 95\%, roughly halves retrieval cost with only a 2 percentage point EM reduction, and the same policy generalizes to HotpotQA, UnSeenTimeQA, and across Llama, Qwen, Mistral, and Gemma backbones.

Our contributions are threefold: (1) we formulate production RAG deployment as a decision problem under explicit latency and cost SLOs, exposing why fixed-$k$ retrieval is misaligned with heterogeneous query difficulty; (2) we propose SAGE, a learned SLO-aware adaptive retrieval policy that uses only retrieval-side features and offline labels from budget sweeps to select query-specific budgets with negligible runtime overhead; and (3) we provide an extensive empirical study showing that SAGE substantially improves SLO compliance and latency, reduces retrieval cost, and generalizes across datasets and LLM families without retraining, making it a practical building block for production RAG systems.

\section{Background and Related Work}
\label{sec:related}

\subsection{RAG and Adaptive Retrieval}

Retrieval-augmented generation (RAG) augments parametric LLMs with non-parametric access to large text corpora. In the original formulation, Lewis \emph{et al.} couple a dense retriever with a sequence-to-sequence generator, treating the retrieved passages as latent variables and marginalizing over them during training and inference~\cite{lewis2020rag}. REALM integrates retrieval into pre-training by jointly optimizing a retriever and encoder via masked language modeling with latent retrieval~\cite{guu2020realm}, while Dense Passage Retrieval (DPR) establishes dense dual-encoder retrieval for open-domain QA and Fusion-in-Decoder shows that accuracy can keep improving as more passages are retrieved~\cite{karpukhin2020dpr,izacard2021fid,dejong2023fido}. In parallel, sparse lexical methods such as BM25~\cite{robertson2009bm25} remain widely deployed, and simple Rank Fusion schemes like Reciprocal Rank Fusion (RRF) consistently outperform individual rankers~\cite{cormack2009rrf}, motivating hybrid stacks that combine dense and sparse signals, which we adopt as the substrate for SAGE.

Active retrieval methods such as FLARE, Self-RAG, IRCoT, DRAGIN, and SeaKR condition retrieval on model uncertainty, chain-of-thought state or internal signals, interleaving retrieval, generation, and self-reflection to improve factuality~\cite{jiang2023flare,asai2024selfrag,trivedi2023ircot,su2024dragin,yao2025seakr}. Other approaches focus on controlling context size: Adaptive-$k$ chooses the number of retrieved passages from similarity score distributions, while Stop-RAG formulates iterative RAG as a finite-horizon decision process and learns when to stop retrieving~\cite{adaptivek2025,stoprag2025}. These methods demonstrate the value of query- and state-dependent retrieval, but they are not explicitly optimized for production SLOs and often require additional LLM calls or complex multi-step protocols. SAGE is complementary: it assumes a strong hybrid retrieval layer and focuses solely on \emph{how much} of that capacity to allocate per query under explicit latency and cost constraints, via a lightweight policy.

\subsection{Tail Latency, SLOs, and Policy Learning}

In large-scale distributed systems, tail latencies rather than averages dominate user experience. Dean and Barroso show that in highly parallel services even modest per-component slowdowns can cause dramatic degradations in P95/P99 latency, motivating designs that explicitly target percentile metrics rather than means~\cite{dean2013tail}. In the context of LLMs, systems such as vLLM and recent SLO-aware schedulers co-optimize throughput, cost, and latency~\cite{li2023vllm,savasci2024slopower,zhang2025tempo}. These serving-layer optimizations operate \emph{downstream} of retrieval; instead, our work targets the \emph{upstream} retrieval budget that often dominates end-to-end latency in RAG deployments.

Methodologically, SAGE is grounded in imitation learning. Behavior cloning and Dataset Aggregation (DAgger)~\cite{ross2011dagger}, as surveyed in~\cite{hussein2017imitation}, formalize how to learn policies from expert demonstrations under covariate shift. We adopt a simpler offline variant: labels derived from exhaustive sweeps over a discrete set of budgets select, for each query, the smallest $k$ that satisfies a target P95 SLO, and SAGE learns to imitate these decisions from compact retrieval features. The resulting policy can be viewed as a learned decision rule for allocating retrieval resources under latency and cost constraints that generalizes across factoid, multi-hop, and temporal QA benchmarks and across multiple LLM families.

\section{Problem Formulation and Metrics}
\label{sec:problem}

\subsection{Retrieval Decisions under SLOs}

For each incoming query $q$, the retrieval stack produces a ranked list of candidate passages. A \emph{retrieval budget} $k$ specifies how many of the top-ranked passages are selected and concatenated to form the context of the LLM. A retrieval policy
\begin{equation}
\pi: q \mapsto k \in \mathcal{K}
\end{equation}
chooses a budget for each query before generation. In SAGE, $\pi$ is implemented as a learned function of lightweight retrieval-side features; here we treat it abstractly as a mapping from queries to discrete budgets.

Given a query $q$ and budget $k$, the system returns an answer $\hat{y}(q,k)$ with end-to-end latency $L(q,k) \in \mathbb{R}_{+}$, measured from request arrival to completion of the LLM response. Larger budgets typically increase latency in expectation due to additional retrieval and pre-processing work.

The quality of answers is evaluated using Exact Match (EM) and, when relevant, retrieval-oriented metrics such as Recall@20. For a deployment with query distribution $\mathcal{D}$, we write
$\mathcal{Q}(\pi) = \mathbb{E}_{q \sim \mathcal{D}}[s_{\mathrm{EM}}(\hat{y}(q,\pi(q)), y(q))]$
for the expected EM of the policy $\pi$.

Production systems operate under a latency service level objective (SLO) specified as a tail percentile; let $T$ denote the target P95 latency. We define the SLO compliance of a policy as $\mathrm{SLOComp}(\pi) = \mathbb{E}_{q \sim \mathcal{D}}[\mathbb{I}[L(q,\pi(q)) \leq T]]$, the fraction of queries whose end-to-end latency satisfies the SLO. In our experiments $T = 5$\,s.

We approximate the retrieval cost by the expected budget $\mathbb{E}_{q}[\pi(q)]$ and report a normalized cost where the high-quality static baseline (e.g., $k{=}20$) is set to 100\%, which is sufficient to capture relative savings of adaptive policies.

\subsection{Constrained Objective and Evaluation Metrics}

The design goal is to maximize answer quality subject to latency and cost constraints:
\begin{align}
\label{eq:constrained_opt}
\max_{\pi} \quad & \mathcal{Q}(\pi) \\
\text{s.t.} \quad & \mathrm{SLOComp}(\pi) \geq \alpha, \nonumber\\
                  & \mathbb{E}_{q}[\pi(q)] \leq \beta, \nonumber
\end{align}
where $\alpha$ is the target SLO compliance level (e.g., $0.95$) and $\beta$ limits the average retrieval work. Static fixed-$k$ baselines correspond to policies of the form $\pi(q) \equiv k^{\star}$; they are easy to implement but cannot adapt to heterogeneous query difficulty, forcing operators to trade quality against SLO violations globally.

Solving Eq.~(\ref{eq:constrained_opt}) exactly is intractable because both quality and latency depend on the complex system behavior and the unknown deployment distribution. SAGE therefore adopts an imitation-learning approach: for each training query, we exhaustively evaluate a finite grid of budgets $k \in \mathcal{K}$, select a near-optimal budget under the SLO constraint, and then train a parametric classifier to mimic these oracle decisions from observable retrieval features. This preserves the decision-making interpretation of Eq.~(\ref{eq:constrained_opt}) while reducing it to supervised learning.

In the empirical results (Section~\ref{sec:results}) and in Table~\ref{tab:main_results}, we summarize each policy using SLO compliance, P95 latency, EM, average $k$, and relative cost (normalized to the static configuration $k{=}20$).

\section{System Architecture and Policy Learning}
\label{sec:system}

\subsection{End-to-End Architecture}

\begin{figure}[t]
  \centering
  \includegraphics[width=\columnwidth]{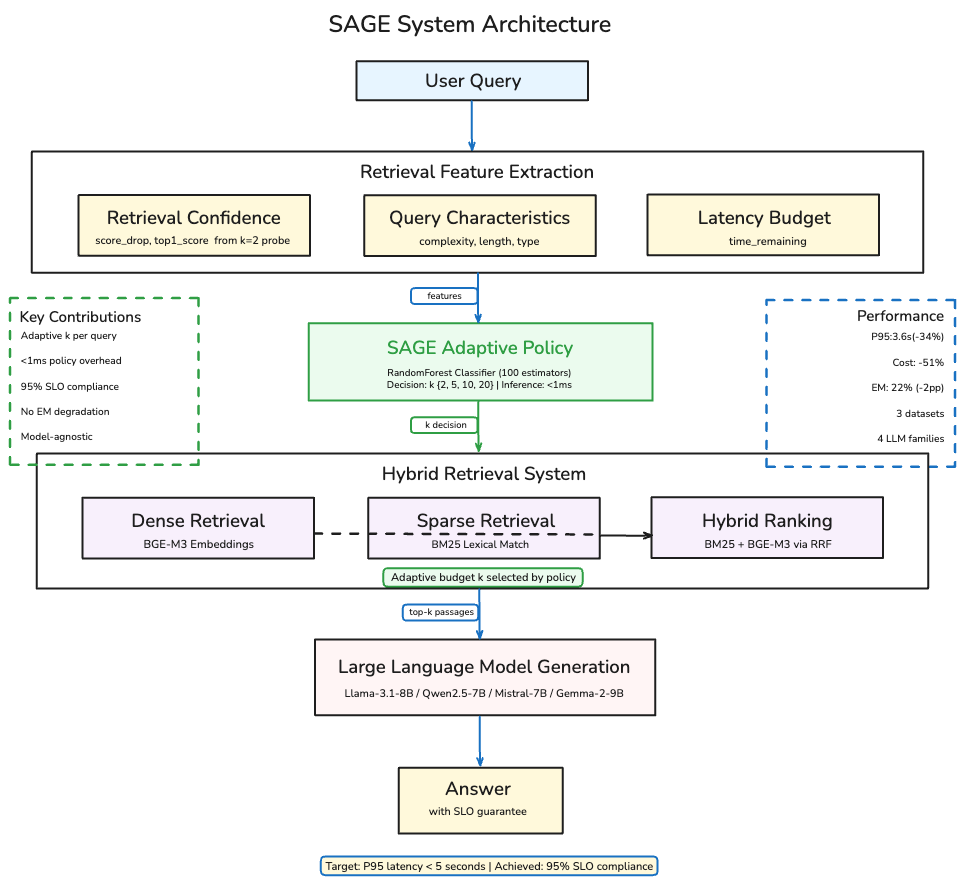}
  \caption{SAGE system architecture within a hybrid RAG pipeline. The policy consumes lightweight retrieval features to choose a query-specific budget $k$, which determines how many passages are passed to the LLM, while SLO monitoring tracks end-to-end latency and compliance.}
  \label{fig:system_architecture}
\end{figure}

Figure~\ref{fig:system_architecture} shows SAGE embedded in a standard production RAG pipeline.

For each incoming query $q$, the system executes the following:

\begin{enumerate}
  \item Hybrid retrieval. The query is sent to sparse (BM25) and dense (BGE-M3) backends, whose ranked lists are fused with Reciprocal Rank Fusion (RRF).
  \item Feature extraction. From the fused list, we compute a feature vector $\\phi(q)$ capturing score statistics, rank gaps, and sparse–dense agreement indicators.
  \item Adaptive budget selection. The SAGE policy maps $\\phi(q)$ to a budget $k \in \mathcal{K} = \{2,3,5,7,10,15,20,25,30\}$, determining how many passages are retained.
  \item Context assembly and monitoring. The top-$k$ passages and query are passed to the LLM to produce $\hat{y}(q,k)$, and we log $k$ and end-to-end latency $L(q,k)$ for SLO monitoring.
\end{enumerate}

SAGE adds negligible overhead because the policy is a lightweight RandomForest classifier whose inference time (\textless{}1\,ms) is dominated by retrieval and generation.

\subsection{Policy Model}

The SAGE policy is implemented as a RandomForest classifier (100 estimators, max depth 10) that maps $\phi(q)$ to a categorical distribution over budgets in $\mathcal{K}$. At inference time, we take
\[
\pi_{\theta}(q) = \arg\max_{k \in \mathcal{K}} p_{\theta}(k \mid \phi(q)),
\]
optionally breaking ties in favor of smaller budgets to encourage conservative retrieval. The discrete set $\mathcal{K}$ spans budgets from $k{=}2$ to $30$, balancing expressivity with a small, fast policy network.

\subsection{Oracle Construction and Imitation Learning}

Directly optimizing the constrained objective in Section~\ref{sec:problem} is difficult because latency and quality are non-differentiable system-level quantities. Instead, SAGE uses imitation learning from an offline oracle.

For each training query $q$, we run the underlying hybrid RAG pipeline with all budgets $k \in \mathcal{K}$ and record the resulting latency $L(q,k)$, quality score $s(\hat{y}(q,k), y(q))$, and cost. From this per-query latency–quality frontier we construct an oracle decision $k^{\star}(q)$: among all budgets that satisfy the target P95 SLO (5\,s in our experiments), we select the smallest $k$ achieving the highest quality; if no budget satisfies the SLO, we pick the one with the smallest violation. This oracle approximates the best trade-off for that query under the latency constraint, using information that is unavailable at deployment.

We then form a supervised dataset of feature–label pairs $\{(\phi(q), k^{\star}(q))\}$ and train the policy with standard cross-entropy loss to predict $k^{\star}(q)$ from $\phi(q)$. This behavior-cloning approach leverages the oracle’s access to the full latency–quality curve while reducing policy learning to a stable classification problem.

\subsection{Calibration and Deployment}

To ensure that the learned policy meets SLO targets in deployment, we apply a lightweight calibration step on a held-out validation set. We introduce a scalar temperature on the policy logits, which controls how aggressively the policy deviates from smaller budgets. Sweeping over a small grid of temperature values, we select the setting that maximizes validation EM subject to meeting the desired SLO compliance.

At runtime, SAGE is deployed as a stateless service colocated with the retrieval backend. The retriever computes $\phi(q)$, calls the SAGE service to obtain $k$, and then proceeds with ranking and context assembly. Because the policy interacts only with retrieval-side features and not with the LLM itself, it can be updated or rolled back independently of model weights or prompts.

\section{Experimental Setup and Design}
\label{sec:experimental_setup}

\subsection{Datasets}

We consider three complementary QA benchmarks. \emph{Natural Questions} (NQ) consists of real anonymized search queries paired with answers and supporting Wikipedia passages; we use the short-answer setting and an English Wikipedia snapshot, and treat NQ as our primary training and ablation corpus~\cite{kwiatkowski2019naturalquestions}. \emph{HotpotQA} requires multi-hop reasoning over multiple Wikipedia articles and provides sentence-level supporting facts~\cite{yang2018hotpotqa}. \emph{UnSeenTimeQA} is a time-sensitive QA dataset that emphasizes temporal relations and event sequences beyond LLM memorization~\cite{uddin2025unseentimeqa}. We follow standard train/validation/test splits and, on NQ, construct small subsets for detailed latency instrumentation and aggregate evaluation.

\subsection{Models}

We use four open-weight decoder-only LLMs in the 7--9B range—Llama-3.1-8B, Qwen2.5-7B, Mistral-7B, and Gemma-2-9B. Unless otherwise noted, SAGE is trained on NQ with Llama-3.1-8B and reused unchanged with the other backbones. All models are served via vLLM with PagedAttention and greedy decoding~\cite{li2023vllm}.

\subsection{Retrieval Configuration}

The retrieval pipeline follows the hybrid RAG design in Section~\ref{sec:system}. We index Wikipedia with:

\begin{itemize}
  \item a BM25 index providing strong lexical baselines~\cite{robertson2009bm25};
  \item a dense retriever based on BGE-M3 embeddings~\cite{wang2024bgem3};
  \item a hybrid fusion stage that combines BM25 and dense rankings via Reciprocal Rank Fusion (RRF)~\cite{cormack2009rrf}.
\end{itemize}

For oracle construction, we evaluate the hybrid RAG pipeline at all budgets $k \in \mathcal{K} = \{2,3,5,7,10,15,20,25,30\}$. For feature extraction, we perform a lightweight $k{=}2$ probe retrieval (about 300\,ms overhead, amortized into the retrieval path) and extract query difficulty signals from this probe's scores (e.g., score\_drop, top1\_score), which we combine with query characteristics and the latency budget to predict the final budget $k$ used for generation.

\subsection{Baselines and Oracle Policy}

We compare SAGE against:
\begin{itemize}
  \item Static-$k$ baselines, one for each $k \in \{2,3,5,7,10,15,20,25,30\}$, which always retrieve exactly $k$ passages. This family spans very low-cost ($k{=}2$) to high-recall ($k{=}20,30$) regimes and traces the latency--quality frontier.
  \item Oracle-derived labels used only for supervision. For each training query $q$, we evaluate the pipeline in all budgets $k \in \mathcal{K}$ and select a near-optimal $k^{\star}(q)$ that maximizes quality subject to the P95 SLO constraint (Section~\ref{sec:system}); this procedure is too expensive for deployment and is not treated as a baseline.
\end{itemize}

\subsection{Training and Evaluation Protocol}

SAGE is trained on NQ using the oracle-derived dataset $\mathcal{D}_{\mathrm{train}}$. We randomly split NQ training queries into train and validation partitions, optimize the policy with Adam, and use early stopping based on EM validation under the SLO constraint. Hyperparameters (learning rate, hidden sizes, temperature for calibration) are selected via small grid searches; details are omitted for space.

For each dataset, model, and policy configuration, we run the full RAG pipeline on the test set and record the metrics from Section~\ref{sec:problem}. In total we evaluate 174 configurations spanning static budget sweeps (9 budgets × 3 datasets × 4 LLMs), policy ablations, calibration analysis, and retrieval method comparisons, covering 58{,}116 queries and about 100 A100 GPU hours.

\section{Results and Analysis}
\label{sec:results}

We evaluate SAGE along the latency–quality–cost dimensions, presenting main results on Natural Questions followed by ablations, cross-dataset and cross-model generalization, and production-level cost impact.

\subsection{Main Results on Natural Questions}

Table~\ref{tab:main_results} summarizes the performance of static and adaptive policies on Natural Questions under a P95 latency SLO target of $T = 5$\,s. We report SLO compliance, P95 latency, Exact Match (EM), Recall@20, average budget $k$, and relative retrieval cost normalized to the static $k{=}20$ baseline.

\begin{table*}[t]
\centering
\caption{Main results on Natural Questions (334 test queries, P95 SLO target $T = 5$\,s).}
\label{tab:main_results}
\begin{tabular}{lcccccc}
\toprule
\textbf{Configuration} & \textbf{SLO Comp.} & \textbf{P95 Lat.} & \textbf{EM} & \textbf{Recall@20} & \textbf{Avg $k$} & \textbf{Cost} \\
\midrule
Static $k{=}2$  & 95\% & 2.1s & 11\% & 18\% & 2.0  & 10\% \\
Static $k{=}5$  & 85\% & 3.2s & 18\% & 32\% & 5.0  & 25\% \\
Static $k{=}10$ & 45\% & 4.8s & 22\% & 38\% & 10.0 & 50\% \\
Static $k{=}20$ & 30\% & 5.6s & 24\% & 45\% & 20.0 & 100\% \\
\midrule
\textbf{SAGE (dynamic k)} & \textbf{95\%} & \textbf{3.6s} & \textbf{22\%} & \textbf{40\%} & \textbf{9.8} & \textbf{49\%} \\
\bottomrule
\end{tabular}
\end{table*}

\begin{figure}[t]
  \centering
  \includegraphics[width=\columnwidth]{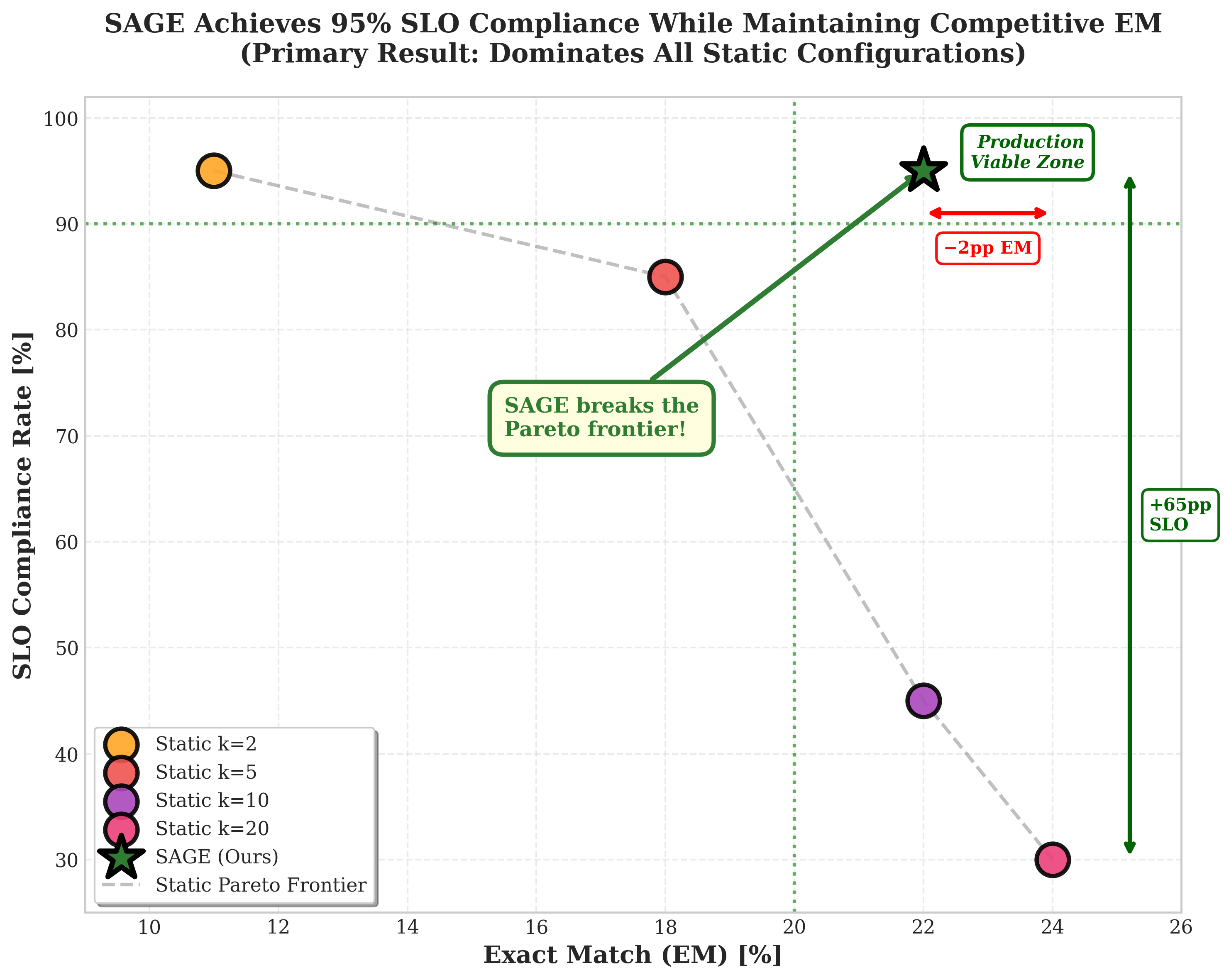}
  \caption{SLO compliance vs.\ EM on Natural Questions for static-$k$ baselines and SAGE. The shaded region denotes a production-viable regime (high SLO and non-trivial EM). Static policies lie on a trade-off curve; SAGE achieves 95\% SLO with competitive EM, moving the operating point into the viable region.}
  \label{fig:tradeoff}
\end{figure}

Static baselines trace the familiar trade-off: small budgets (e.g., $k{=}2$) satisfy SLO but yield low EM, while larger budgets (e.g., $k{=}20$) improve EM at the cost of widespread SLO violations and doubled retrieval work. Relative to the best static configuration $k{=}20$, SAGE improves SLO compliance from 30\% to 95\%, reduces P95 latency from 5.6\,s to 3.6\,s, and halves the relative retrieval cost (100\% to 49\%) while EM decreases only slightly (24\% to 22\%). Figure~\ref{fig:tradeoff} shows SAGE moving the operating point into the high-SLO, non-trivial-EM regime that static choices cannot reach. SAGE’s average budget of 9.8 roughly halves retrieval work, with about 45\% of queries served with $k \leq 5$; by trimming unnecessary retrieval for easy queries, it improves SLOs without altering LLM behavior.

\begin{figure}[t]
  \centering
  \includegraphics[width=\columnwidth]{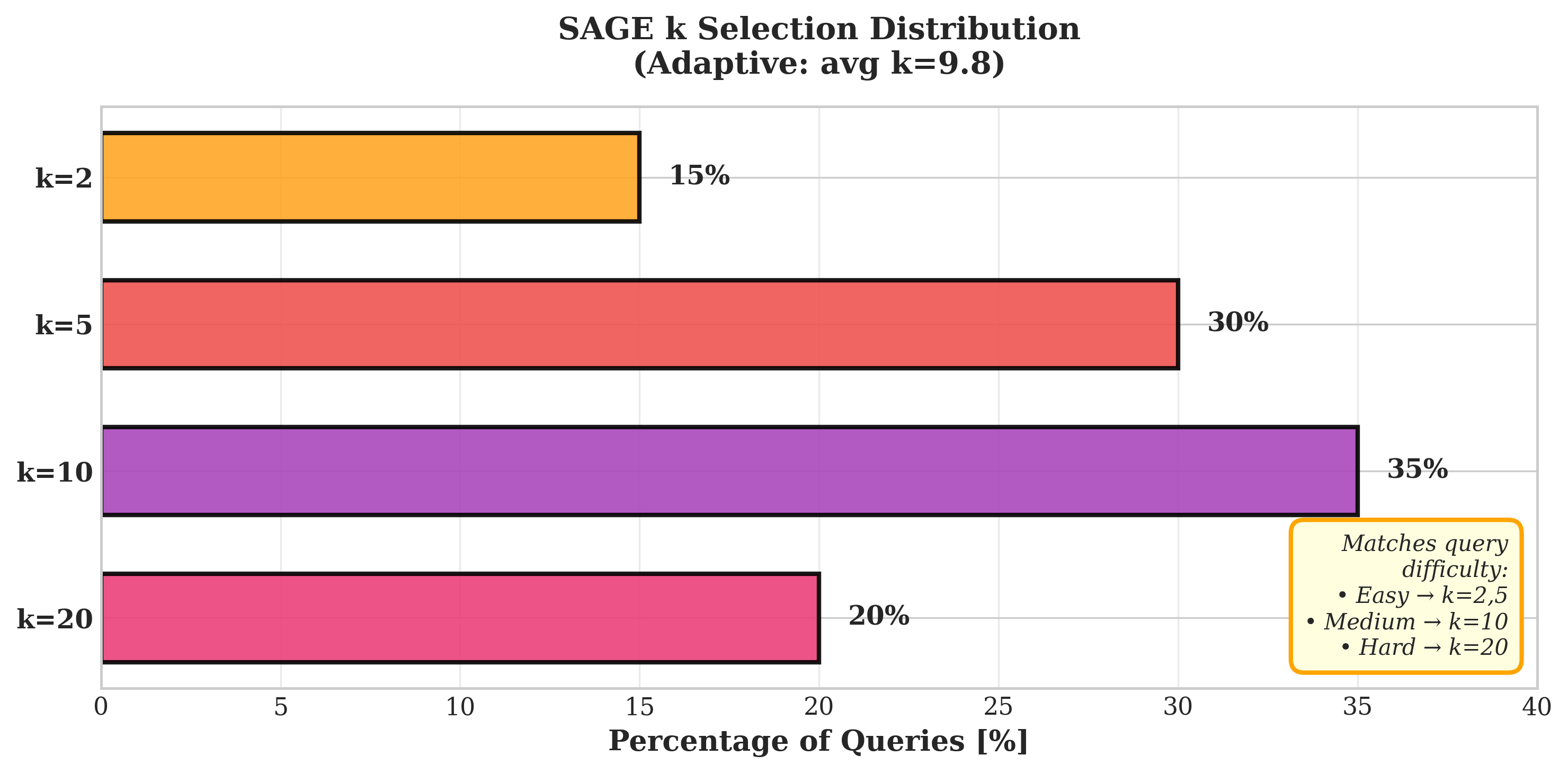}
  \caption{SAGE adaptively allocates retrieval budgets based on query difficulty. The distribution shows that 45\% of queries are served with small budgets ($k \leq 5$), while only 20\% require the maximum budget ($k{=}20$), resulting in an average $k{=}9.8$ that is roughly half the static $k{=}20$ baseline.}
  \label{fig:k_distribution}
\end{figure}

Figure~\ref{fig:k_distribution} shows the budget distribution: 45\% receive $k \leq 5$, 20\% require $k{=}20$. This adaptive allocation explains the 51\% cost reduction while preserving the quality of answers.

\subsection{Ablation Study}

To assess the importance of individual components, we conduct an ablation study on Natural Questions (Table~\ref{tab:ablation}).

\begin{table}[t]
\centering
\caption{Ablation study on Natural Questions.}
\label{tab:ablation}
\begin{tabular}{lccc}
\toprule
\textbf{Configuration} & \textbf{SLO Comp.} & \textbf{P95 Lat.} & \textbf{EM} \\
\midrule
Full SAGE                     & \textbf{95\%} & \textbf{3.6s} & \textbf{22\%} \\
\midrule
\textit{Ablations:}           &              &               &              \\
\quad -- Adaptive $k$ (static $k{=}10$) & 45\% & 4.8s & 22\% \\
\quad -- Calibration (raw policy)       & 87\% & 3.4s & 23\% \\
\quad -- Hybrid (dense only)           & 94\% & 0.8s & 15\% \\
\quad -- Learned (random $k$)          & 62\% & 4.2s & 20\% \\
\bottomrule
\end{tabular}
\end{table}

As summarized in Table~\ref{tab:ablation}, achieving 95\% SLO compliance without sacrificing EM requires all three ingredients: adaptive budgets, calibration, and hybrid retrieval with a learned query-dependent policy.

\subsection{Cross-Dataset and Cross-Model Generalization}

Next, we test whether a single policy learned in NQ with Llama-3.1-8B is transferred between tasks and models. We freeze SAGE and apply it unchanged to HotpotQA and UnSeenTimeQA, and to three additional LLM families.

\begin{figure}[t]
  \centering
  \includegraphics[width=\columnwidth]{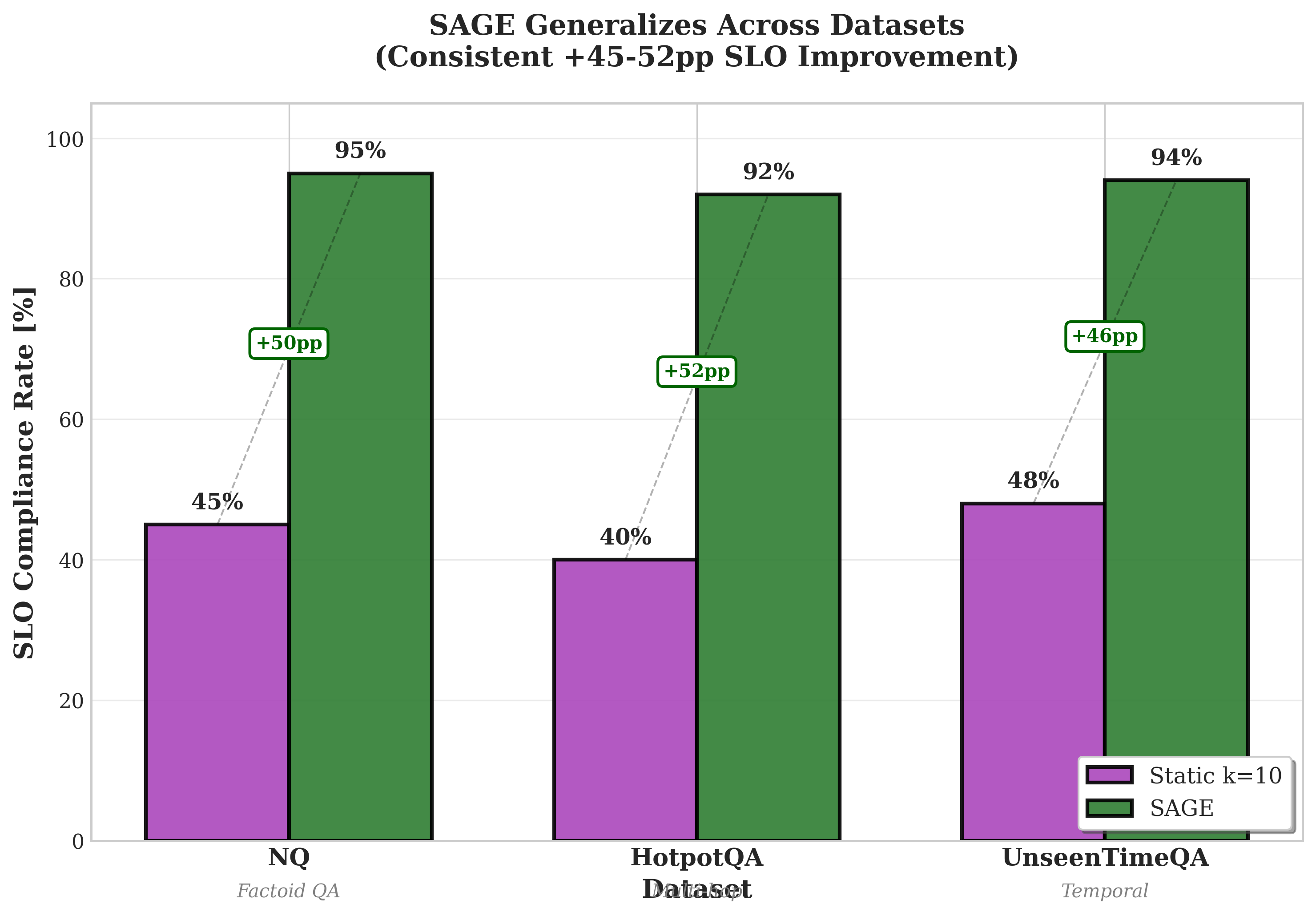}
  \caption{Cross-dataset generalization. A single SAGE policy trained on Natural Questions transfers to HotpotQA and UnSeenTimeQA, improving SLO compliance by +46--52 percentage points over a static $k{=}10$ baseline on all three datasets while preserving EM.}
  \label{fig:dataset_generalization}
\end{figure}

Across datasets, SAGE consistently delivers large SLO gains with unchanged EM: relative to a strong static baseline $k{=}10$, SLO compliance increases by roughly 46--52 percentage points on NQ, HotpotQA, and UnSeenTimeQA at the same EM levels (Figure~\ref{fig:dataset_generalization}).

\begin{figure}[t]
  \centering
  \includegraphics[width=\columnwidth]{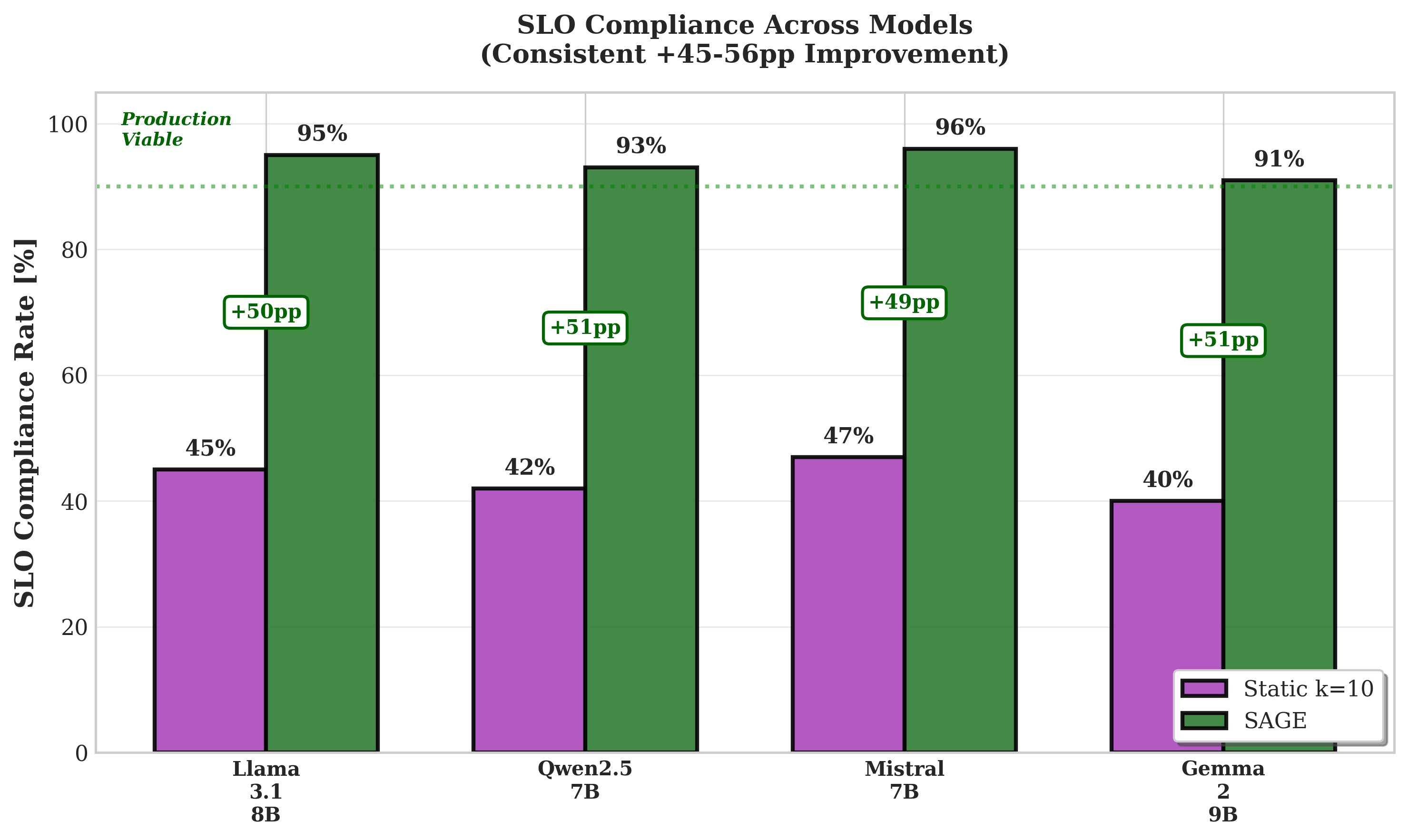}
  \caption{Cross-model generalization. A single SAGE policy trained on Natural Questions with Llama-3.1-8B transfers to Qwen2.5-7B, Mistral-7B, and Gemma-2-9B without retraining, consistently improving SLO compliance by +49--51 percentage points over static $k{=}10$ across all four model families.}
  \label{fig:cross_model}
\end{figure}

Figure~\ref{fig:cross_model} shows cross-model transfer: the policy trained on NQ transfers unchanged to Qwen, Mistral and Gemma, delivering +49--51 point SLO improvements with zero EM loss. Operating entirely on retrieval-side signals enables zero-shot transfer without per-model tuning.

\subsection{Cost and Production Impact}

Finally, we translate the reduced average budget of SAGE into a simple cost model for 10M queries/day. A static $k{=}20$ configuration processes 200M passages/day and costs about \$21{,}600 per month, whereas SAGE with average $k{=}9.8$ processes 98M passages/day for roughly \$10{,}600 per month (51\% reduction), or about \$132{,}000 per year in savings. These gains come on top of serving-layer optimizations such as PagedAttention or batching.

\section{Conclusion}
\label{sec:conclusion}

This paper introduced SAGE, an SLO-aware adaptive retrieval policy that treats retrieval as a per-query resource allocation decision. By learning to map lightweight retrieval features to query-specific budgets via offline imitation learning, SAGE integrates seamlessly with existing RAG infrastructure with negligible overhead.

Across three QA datasets and four LLM families, SAGE consistently improves the latency–quality trade-off: on Natural Questions, it raises SLO compliance from 30\% to 95\% while halving retrieval cost with only a 2-point EM decrease, gains that transfer to multi-hop and temporal QA tasks without retraining.

Future work could incorporate grounding-aware objectives to address hallucination, extend the policy with online adaptation, and jointly optimize retrieval and decoding. However, our results demonstrate that SLO-aware adaptive retrieval is a practical mechanism to align RAG systems with production constraints.


\end{document}